%% file: sample-sigconf.tex
\documentclass[sigconf]{acmart}

\usepackage{kotex}
\usepackage{cleveref}
\usepackage{amsmath}
\usepackage{algorithm}
\usepackage{algorithmicx}
\usepackage{algpseudocode}
\usepackage{booktabs}
\usepackage{multirow}
\usepackage{tabularx}
\usepackage{optidef}
\usepackage{breqn}
\usepackage{dsfont}
\usepackage{subfig}
\usepackage{url}
\usepackage{pifont}
\newcommand{\vmark}{\ding{51}}
\newcommand{\xmark}{\ding{55}}

\AtBeginDocument{%
  \providecommand\BibTeX{{%
    \normalfont B\kern-0.5em{\scshape i\kern-0.25em b}\kern-0.8em\TeX}}}

\copyrightyear{2023}
\acmYear{2023}
\setcopyright{acmlicensed}
\acmConference[WSDM '23] {Proceedings of the Sixteenth ACM International Conference on Web Search and Data Mining}{February 27--March 3, 2023}{Singapore, Singapore.}
\acmBooktitle{Proceedings of the Sixteenth ACM International Conference on Web Search and Data Mining (WSDM '23), February 27--March 3, 2023, Singapore, Singapore}
\acmPrice{15.00}
\acmISBN{978-1-4503-9407-9/23/02}
\acmDOI{10.1145/3539597.3570439}

\makeatletter
\def\@fnsymbol#1{\ensuremath{\ifcase#1\or \dagger\or *\or \ddagger\or
   \mathsection\or \mathparagraph\or \|\or **\or \dagger\dagger
   \or \ddagger\ddagger \else\@ctrerr\fi}}

\begin{document}

\title{Reliable Decision from Multiple Subtasks through Threshold Optimization: Content Moderation in the Wild}


\author{Donghyun Son}
\authornote{Equal contribution.}
\email{ryan.s@hpcnt.com}
\affiliation{%
  \institution{Hyperconnect}
  \city{Seoul}
  \country{South Korea}
}

\author{Byounggyu Lew}
\authornotemark[1]
\email{korts@hpcnt.com}
\affiliation{%
  \institution{Hyperconnect}
  \city{Seoul}
  \country{South Korea}
}

\author{Kwanghee Choi}
\authornotemark[1]
\email{kwanghee.choi@hpcnt.com}
\affiliation{%
  \institution{Hyperconnect}
  \city{Seoul}
  \country{South Korea}
}

\author{Yongsu Baek}
\email{hunter@hpcnt.com}
\affiliation{%
  \institution{Hyperconnect}
  \city{Seoul}
  \country{South Korea}
}

\author{Seungwoo Choi}
\email{seungwoo.choi@hpcnt.com}
\affiliation{%
  \institution{Hyperconnect}
  \city{Seoul}
  \country{South Korea}
}

\author{Beomjun Shin}
\email{beomjun.shin@match.com}
\affiliation{%
  \institution{Match Group}
  \city{Dallas, Texas}
  \country{USA}
}

\author{Sungjoo Ha}
\email{sungjoo.ha@hpcnt.com}
\affiliation{%
  \institution{Hyperconnect}
  \city{Seoul}
  \country{South Korea}
}

\author{Buru Chang}
\authornote{Corresponding author.}
\email{buru.chang@hpcnt.com}
\affiliation{%
  \institution{Hyperconnect}
  \city{Seoul}
  \country{South Korea}
}
\renewcommand{\shortauthors}{Donghyun Son et al.}

\input{sections/0_abstract}

\maketitle

\sloppy
\input{sections/1_introduction}

\input{sections/2_multiple_subtask_approach}
\input{sections/3_threhold_optimization}
\input{sections/4_experiments}
\input{sections/5_analysis}
\input{sections/6_related_work}
\input{sections/7_conclusion}


\bibliographystyle{ACM-Reference-Format}
\bibliography{sample-base}

\end{document}

%% file: sections/0_abstract.tex
\begin{abstract}
Social media platforms struggle to protect users from harmful content through content moderation. 
These platforms have recently leveraged machine learning models to cope with the vast amount of user-generated content daily.
Since moderation policies vary depending on countries and types of products, it is common to train and deploy the models per policy.
However, this approach is highly inefficient, especially when the policies change, requiring dataset re-labeling and model re-training on the shifted data distribution. 
To alleviate this cost inefficiency, social media platforms often employ third-party content moderation services that provide prediction scores of multiple subtasks, such as predicting the existence of underage personnel, rude gestures, or weapons, instead of directly providing final moderation decisions.
However, making a reliable automated moderation decision from the prediction scores of the multiple subtasks for a specific target policy has not been widely explored yet.
In this study, we formulate real-world scenarios of content moderation and introduce a simple yet effective threshold optimization method that searches the optimal thresholds of the multiple subtasks to make a reliable moderation decision in a cost-effective way.
Extensive experiments demonstrate that our approach shows better performance in content moderation compared to existing threshold optimization methods and heuristics.
\end{abstract}

\ccsdesc[500]{Computing methodologies~Machine learning approaches}
\ccsdesc[500]{Social and professional topics~Censorship}

\keywords{Content Moderation, Threshold Optimization, Multiple Subtasks}

%% file: sections/1_introduction.tex
\section{Introduction}\label{sec:1_introduction}
\input{figures/1_multiple_subtask_example}

Users of social media platforms that provide user-generated content are always at risk of being exposed to harmful content.
To protect the users, most social media platforms operate content moderation systems with human moderators to classify whether a given content is problematic or not \cite{gorwa2020algorithmic}.
However, human decisions are often noisy~\cite{yun2021re}, and a vast amount of content are generated daily, so it is challenging to handle content only by manual labor.
Furthermore, caring for moderators' mental health while handling harmful content is becoming an important issue~\cite{karunakaran2019testing}.

To alleviate the above issues, many social media platforms (e.g., Facebook and Instagram) have utilized machine learning (ML) models to automatically review a large quantity of content without human moderators.
However, users can still be exposed to harmful content due to the ML model's imperfectness.
This risk imposes a substantial burden on social media platforms under tighter government regulations on inappropriate content \cite{gorwa2020algorithmic}.
To mitigate this, these platforms integrate the capabilities of both humans and machines: trust the decision of ML models only when their confidence is sufficiently high, while the human moderators handle the rest.
Hence, the ML model must make a reliable moderation decision, at least on par with its human counterparts, to reduce the total volume of content to be handled by the moderators.

To improve the quality of automated decisions, the ML model performance should increase, thus requiring a large-scale dataset.
However, it is challenging because the same content may be annotated differently depending on moderation policies varying with country or product.
For this reason, the social media platforms need to construct and manage datasets for each moderation policy separately to train and deploy an individual model for the corresponding policy.
Furthermore, this approach gradually increases the labeling cost as the moderation policies change due to everchanging circumstances such as new legislation or societal demand.
The existing dataset should be re-labeled under the changing policies, and the pretrained model should also be re-trained on the shifted data distribution~\cite{wang2018deep}.

Some social media platforms often employ third-party content moderation services (e.g., Hive moderation\footnote{https://docs.thehive.ai/} and Azure content moderator\footnote{https://azure.microsoft.com/en-us/services/cognitive-services/content-moderator/}) to avoid aforementioned issues.
As shown in Figure~\ref{fig:1_multiple_subtask_example}, these services provide prediction scores of the multiple subtasks, such as predicting the existence of weapons and rude gestures in a given content, instead of directly providing moderation decisions tailored for each specific service.
Customers (social media platforms) make decisions from the provided prediction scores by utilizing or ignoring provided the scores, i.e., defining the decision function (refer to Section~\ref{subsec:2_3_decision_function} for the definition).
Customers can easily cope with the changing policies by modifying the decision function.
To effectively support the customers, moderation services have to define each subtask to be granular, exhaustively covering the potential moderation policies that customers might take.

Although this multiple subtask approach relieves the customers' burden to operate their own ML models and maintain private datasets, it makes utilizing the prediction scores more complicated.
Customers have to first apply thresholding to the prediction scores to a binary form (e.g., exist or not).
Then, they design the decision function that digests the thresholded multiple subtask predictions to yield a single moderation decision.
However, since prediction scores of ML models are often not calibrated \cite{guo2017calibration}, it is difficult to determine the optimal thresholds for each subtask that maximize the target metrics (e.g., recall at precision).
For example, $0.42$ could be a high enough threshold for subtask A but not for subtask B.
Hence, when we use the same threshold for all the subtask predictions, performance is suboptimal, failing to have effective yet reliable automated decisions.


In this paper, we claim that the sophisticated decision function with the optimal thresholds for prediction scores of multiple subtasks can further improve moderation performance.
To do this, we first formulate real-world content moderation scenarios and summarize the concept of the multiple subtask approach.
We then propose a simple yet effective threshold optimization method that determines the optimal thresholds of multiple subtasks within a few seconds to maximize the number of automated decisions while maintaining their reliability (i.e., recall at precision).
Through experiments on synthetic and real-world moderation datasets, we show that our proposed method outperforms the baselines, including heuristics designed by moderation experts.

\textbf{Contributions.} (1) We introduce real-world content moderation scenarios through the multiple subtask approach to cope with changing moderation policies.
(2) We propose an efficient threshold optimization method to make moderation decisions more reliable in the multiple subtask approach where thresholds are found within a few seconds.
(3) Our proposed method outperforms existing baselines in both synthetic and real-world moderation datasets.

%% file: figures/1_multiple_subtask_example.tex
\begin{figure*}[t] 
\begin{center}
\includegraphics[width=0.90\textwidth]{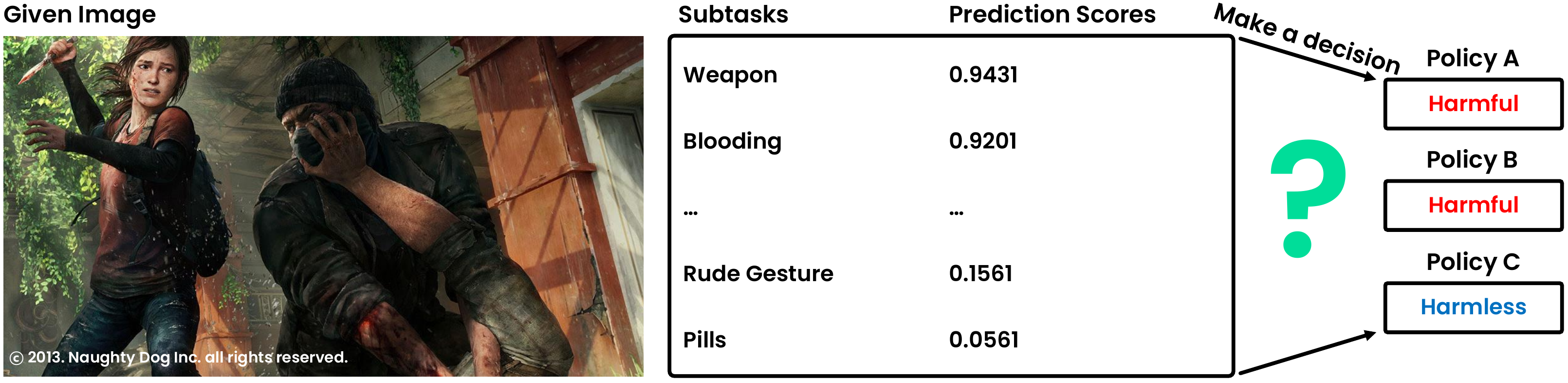}
\end{center}
\caption{An example of the multiple subtask approach.
Instead of providing moderation decisions directly, content moderation services give the prediction scores of the multiple subtasks.
Their customers (social media platforms) rely heavily on heuristic decision functions designed for every policy to make a reliable moderation decision based on the given prediction scores.}
\label{fig:1_multiple_subtask_example}
\end{figure*}

%% file: sections/2_multiple_subtask_approach.tex
\section{Multiple Subtask Approach}\label{sec:2_multiple_subtask_approach}

\subsection{Content Moderation}\label{subsec:2_1_content_moderation}
Content moderation $f$ is commonly defined as classifying whether a given content $x$ is harmless (\textit{class 0}) or harmful (\textit{class 1}) under a target policy $p$:
\begin{equation}
    f_p(x) = 
        \begin{cases}
            1, & \text{if $x$ is harmful,}  \\
            0, & \text{otherwise.}
        \end{cases}
\label{eq:common_content_moderation}
\end{equation}

Although the performance of ML models has been dramatically improved, considering the risk of missing harmful content, the performance is yet to be perfect to automatically moderate all the content based on the ML models.
Therefore, in the real-world, content moderation is designed with the collaboration of the ML models and human moderators to reduce the risk of missing harmful content~\cite{green2019principles,lubars2019ask}.
Considering this real-world scenario, we formulate ML model-based content moderation.

As shown in Figure~\ref{fig:2_content_moderation_use_cases}, classes $\{0,1\}$ of ML-based content moderation are defined differently depending on the use cases.

\input{figures/2_content_moderation}

\noindent
\textbf{Use case 1.} When content moderation targets reported content by users, the reported content should be reviewed because it would be harmful with a high probability.
In this case, an ML model $m_p$ is utilized to automatically filter out harmful content to reduce the number of manual reviews.
\begin{equation}
    f_p(x) = 
        \begin{cases}
            1, & \text{if $m_p(x) > \tau_p$,}  \\
            0, & \text{otherwise,}
        \end{cases}
\label{eq:use_case_1}
\end{equation}
where $m_p(x)$ is the predicted probability by the ML model $m_p$ on whether the given content $x$ is harmful.
$\tau_p$ is a policy-dependent threshold determining whether to automatically filter out the given content (\textit{class 1}) or hand it over to human moderators (\textit{class 0}).

\noindent
\textbf{Use case 2.} When reviewing newly generated content, the human moderators cannot handle the vast amount of all the content.
In this case, an ML model $m_p$ is utilized to automatically skip reviews on a large amount of clearly harmless content:
\begin{equation}
    f_p(x) = 
        \begin{cases}
            1, & \text{if $m_p(x) > \tau_p'$,}  \\
            0, & \text{otherwise,}
        \end{cases}
\label{eq:use_case_2}
\end{equation}
where $m_p(x)$ is the predicted probability of the given content $x$ being harmless.
$\tau_p'$ is a threshold of the target policy $p$ determining whether to automatically skip inspection (\textit{class 1}) or review the content manually (\textit{class 0}).

\subsection{Multiple Subtask Approach}\label{subsec:2_2_multiple_subtasks}
As described in previous sections, ML models $m_p$ are often trained and deployed per policy $p$.
However, this approach becomes impractical when handling multiple changing policies due to service-specific circumstances, requiring data re-labeling and model re-training.
To address this issue, some social media platforms or third-party content moderation services take the multiple subtask approach instead of operating policy-dependent ML models, as shown in~Figure~\ref{fig:3_multiple_task_approach}.

The multiple subtask approach starts with defining $n$ subtasks that are granular enough to cover all the moderation scenarios.
For example, one can subdivide hate speech into ethical, racial, and sexual violence words for text content moderation~\cite{de2018hate}.
Since the subtasks focus on detecting highly specified information in the given content rather than policy-dependent moderation decisions, it allows us to continuously accumulate data for each subtask even if policies change.
It naturally leads to performance improvement of the ML model targeting a specific subtask.

Let $S=\{s_1, s_2, \cdots, s_n\}$ denote a set of subtasks where each subtask $s_i$ is defined as a binary classification for a given content $x$:
\begin{equation}
    s_i(x) = 
        \begin{cases}
            1, & \text{if $m_{s_i}(x)>\tau_{s_i}$ ,}  \\
            0, & \text{otherwise,}
        \end{cases}
\label{eq:subtask}
\end{equation}
where $m_{s_i}(x)$ is the subtask-specific prediction of the ML model $m_{s_i}$ and $\tau_{s_i}$ is the threshold for the subtask.
The output of the multiple subtask models is denoted as $O_S(x) = \{s_1(x), s_2(x), \cdots, s_n(x)\}$ = $\{0,1\}^{n}$.
Either $n$ subtask-specific models \cite{houlsby2019parameter} or a single multi-task model~\cite{ruder2017overview} can be used to implement the above.

\input{figures/3_multiple_subtask_approach}

\subsection{Decision Function}\label{subsec:2_3_decision_function}
Using the multiple subtask output $O_S(x)$, the final moderation decision $f_p(x)$ for the given content $x$ is made with a decision function designed for the policy $p$.
The decision function is manually designed as logical operations applied to the output $O_S(x)$ to fulfill the target policy $p$.

Let $S_p$=$\{s^p_1, s^p_2, \cdots\}$ denote a set of the subtasks $s^p \in S$ which are ground of moderation decision under the target policy $p$.
Using the output $O_{S_p}(x)=\{s^p_1(x), s^p_2(x), \cdots\}$, the final moderation decision $f_p(x)$ is made with the decision function $d_p(\cdot) \rightarrow \{0,1\}$ as follows:
\begin{equation}
    \begin{split}
    f_p(x)&= d_p(O_{S_p}(x)) = d_p\Big(\Big\{(m_{s^p}(x)>\tau_{s^p})\Big\}_{s^p \in S_p}\Big).
    \end{split}
\label{eq:decision_function} 
\end{equation}

For a better understanding, let's assume the situation of content moderation: social media platform for children.
The platform should maintain a policy to remove violent images harmful to children.
Then, the decision function $d_p$ is defined with boolean operations \texttt{AND} and \texttt{OR} on the three subtasks, existence of kids $s_{\text{kids}}$, weapons $s_{\text{weapon}}$, and physical violence $s_{\text{violence}}$, as follows.
\begin{equation}
\begin{split}
    d_p(O_{S_p}(x)) = \Big(s_{\text{kids}}(x) \; &\texttt{AND} \Big(s_{\text{weapon}}(x) \; \texttt{OR} \; s_{\text{violence}}(x)\Big)\Big).
\end{split}
\label{eq:example_of_decision_function}
\end{equation}

%% file: figures/2_content_moderation.tex
\begin{figure}[t] 
\begin{center}
\includegraphics[width=0.90\columnwidth]{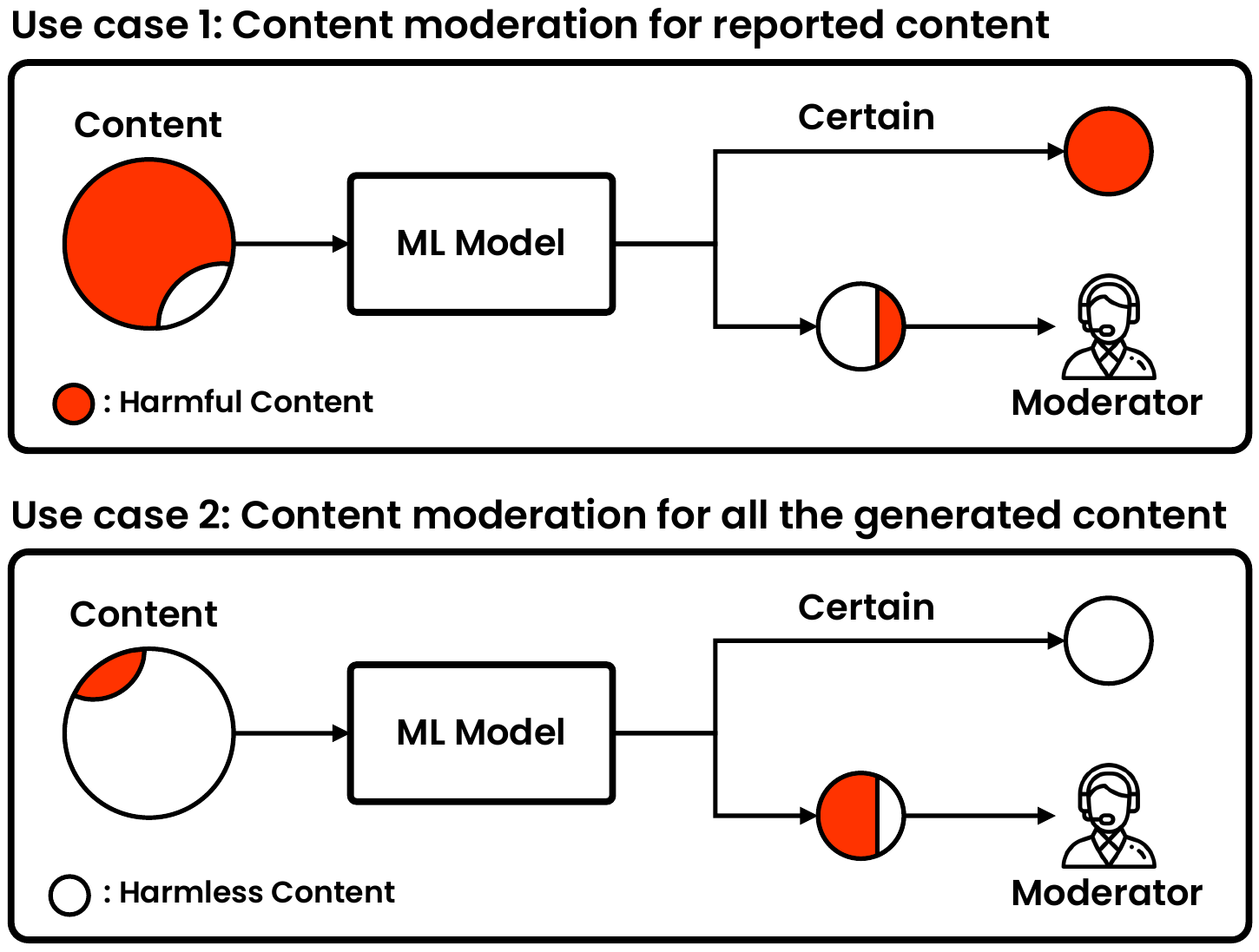}
\end{center}
\caption{Two content moderation use cases of ML models.
Due to the risk of missing harmful content, moderation decisions are made with the aid of human moderators rather than relying solely on model predictions.}
\vspace*{-1.0em}
\label{fig:2_content_moderation_use_cases}
\end{figure}

%% file: figures/3_multiple_subtask_approach.tex
\begin{figure}[t] 
\begin{center}
\includegraphics[width=0.9\columnwidth]{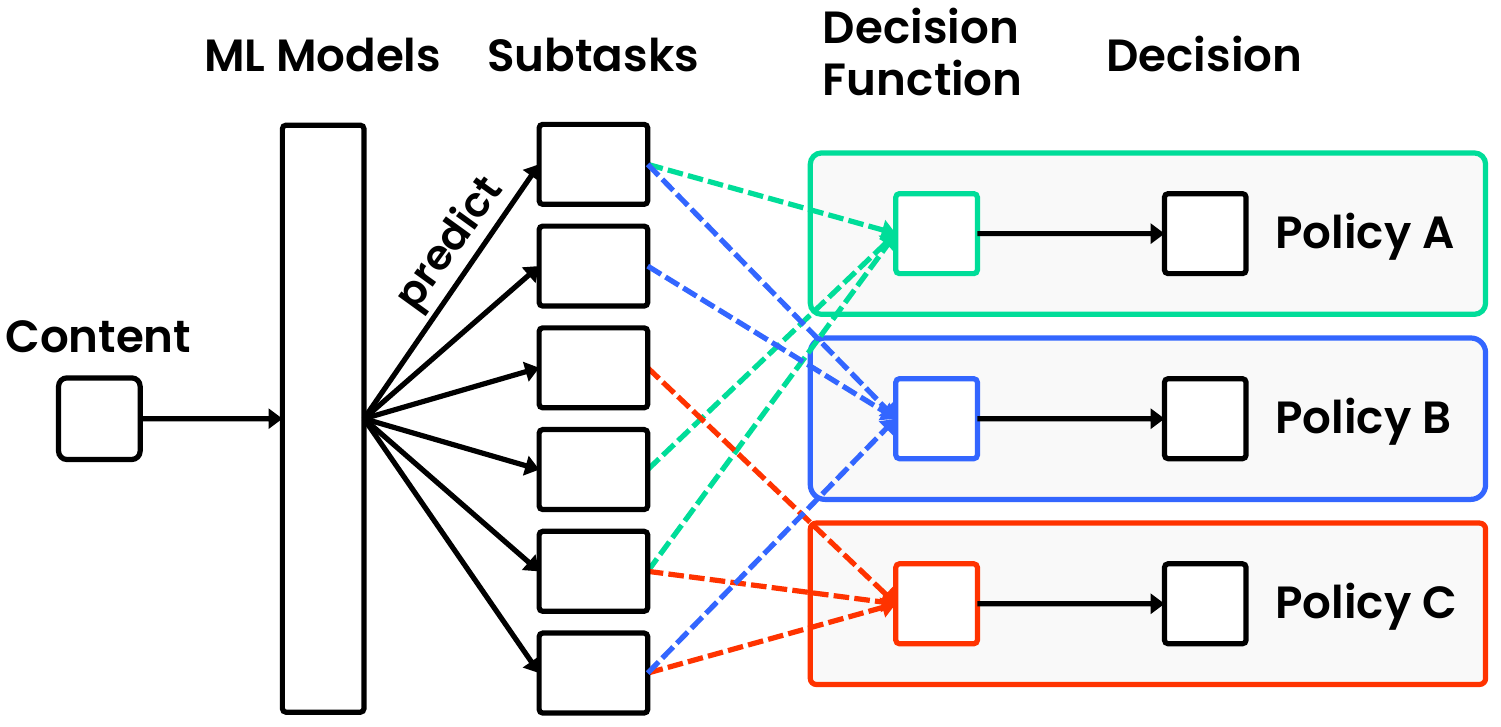}
\end{center}
\caption{
The multiple subtask approach. Instead of making moderation decisions from model predictions directly, decisions are made from the prediction scores of multiple subtasks.
Each subtask covers various policies, where a decision function is designed for each specific target policy.
}
\label{fig:3_multiple_task_approach}
\end{figure}

%% file: sections/3_threhold_optimization.tex
\section{Threshold Optimization}\label{sec:3_threshold_optimization}
As described in the use cases of ML models in content moderation (Figure~\ref{fig:2_content_moderation_use_cases}), ML model predictions should be reliable enough to skip manual reviews.
When ML models make wrong moderation decisions, we incorrectly restrict harmless content (\textit{in Use case 1}) or expose harmful content to users (\textit{in Use case 2}).
Hence, it is crucial to make ML model-based moderation decisions more reliable.

What makes moderation decisions reliable?
It is known that higher model prediction scores are more reliable, i.e., show higher precision \cite{hendrycks2016baseline}.
From this, one may set human precision as the lower bound for automated moderation decisions so that the model is at least not worse than its human counterparts.
On the other hand, increasing the recall is also important to lower the moderation cost.
As precision and recall have a trade-off relationship, we target to maximize recall at high precision to reduce the moderation cost while preserving the reliability of automated decisions.

There are two possible approaches to improve the final moderation decision in Equation~\ref{eq:decision_function}.
One way is to improve the quality of model predictions $m_{s^p}(x)$, which can be challenging in many cases.
Another is to optimize the thresholds $\tau_s$ for each subtask, balancing them to obtain the optimal decision.
We focus on the latter, a much cheaper way to solve the problem than the former.

\subsection{Problem Definition}\label{subsec:3_1_problem_definition}
Let $p$ be the target policy and $T_p = \{{\tau}^{p}_1, {\tau}^{p}_2, \cdots\}$ be the thresholds for the multiple subtasks $S_p=\{s^p_1, s^p_2, \cdots\}$.
We formulate the problem of interest as follows:
\begin{equation}
    \begin{split}
        \underset{T_p}{\text{maximize}}\  &\text{recall}(T_p | \mathcal{D}_p , d_p) \\
        \text{subject to } &\text{precision}(T_p | \mathcal{D}_p , d_p) \geq \text{precision}_t.
    \end{split}
\end{equation}
$\mathcal{D}_p$ is a dataset collected for the policy $p$, where each data sample consists of the prediction scores $Q_j = \{q_{0j}, q_{1j}, \cdots\}$ for the given content $x_j$ and the ground truth decision $y^p_j$ of content moderation.
For simplicity, we refer $m_{s^p_i}(x_j)$ as $q_{ij}$ in the remainder of this paper.
$\text{recall}(T_p | \mathcal{D}_p, d_p)$ and $\text{precision}(T_p | \mathcal{D}_p, d_p)$ are calculated on the dataset $\mathcal{D}_p$ for the given thresholds $T_p$ and the decision function $d_p$ designed for the target policy $p$.
Moderation providers target $\text{precision}_t$ to ensure the reliability of ML-based content moderation.

\subsection{Proposed Method}\label{subsec:3_2_our_approach}
We introduce our threshold optimization method called \textbf{TruSThresh} (\textbf{Tru}ncated \textbf{S}urrogate gradient for \textbf{Thresh}old optimization).
TruSThresh involves 1) surrogate gradient learning (SGL) \cite{ neftci2019surrogate} with width truncation to efficiently tune the thresholds, and 2) a penalty method to meet the target precision.

\noindent
\subsubsection{Score Normalization.}\label{subsubsec:3_2_1_normalization} 
In real-world scenarios, prediction score distributions are often skewed as each subtask facilitates different data distribution.
This skewness makes gradient-based threshold optimization much harder because gradients generated by the prediction scores would also be skewed. 
Thus, we normalize the prediction scores $q_{ij}$ per each subtask $s_i$ as follows:
\begin{equation}
    {\hat{q}_{ij}} = rank(q_{ij}, \{q_{i1}, q_{i2}, \cdots, q_{i|\mathcal{D}_p|}\}) / {|\mathcal{D}_p|},
\label{eq:normalization} 
\end{equation}
where $|\mathcal{D}_p|$ is the number of samples in $\mathcal{D}_p$ and rank function $rank(x, X)$ returns the rank of $x$ in the set $X$ sorted in ascending order.
The rank of $q_{ij}$ is divided by $|\mathcal{D}_p|$ so that the normalized score $\hat{q}_{ij}$ lies between $0$ and $1$. 
Through normalization, we make our optimization strategy more robust by making it agnostic to the prediction score distributions.

We modify Equation~\ref{eq:subtask} with the normalized prediction scores to define normalized subtasks $\hat{S}_p = \{\hat{s}^p_0, \hat{s}^p_1, \cdots\}$ parameterized by $\hat{T}_p = \{\hat{\tau}^p_0, \hat{\tau}^p_1, \cdots\}$ as follows:
\begin{equation}
    \hat{s}_i(x_j) = \begin{cases}
            1, & \text{if $\hat{q}_{ij} > \hat{\tau}^{p}_{i}$,}  \\
            0, & \text{otherwise,}
        \end{cases}
\end{equation}
where $\hat{\tau}^p_i$ is a threshold for the normalized subtask $\hat{s}^p_i$.
Since the normalization does not affect the order of scores, we can easily convert $\hat{T}_p$ back to ${T}_p$ using linear interpolation.

\noindent
\subsubsection{Surrogate Gradient Learning with Width Truncation.}\label{subsubsec:3_2_2_surrogate_gradient_learning}
Similar to \citet{pellegrini2021fast}, our threshold optimization procedure is mainly based on SGL that learns parameters through back-propagation.
The backward pass of SGL is computed using the surrogate gradient, while the forward pass is computed using the step function.
\citet{pellegrini2021fast} use the derivative of the sigmoid function as a surrogate gradient while using the Heaviside step function (HSF) in the forward pass. 
In this work, we also use HSF for the forward pass, but for the backward pass, we approximate the step function using the sine function.

\input{figures/4_method}

\noindent \textbf{Forward Pass.} In the forward pass, we first make binary prediction $\tilde{y}^p$ with $\hat{T}^p$ to compute precision, recall, and loss. 
The final binary prediction $\tilde{y}^p$ for $f_p$ is as follows:
\begin{equation}
\begin{split}
\tilde{y}^p &= \tilde{d}_p\Big(\Big\{\hat{s}^p_0(x), \hat{s}^p_1(x), \cdots\Big\}\Big) \\
            &= \tilde{d}_p\Big(\Big\{\text{HSF}(\hat{q}_0 - \hat{\tau}^{p}_0), \text{HSF}(\hat{q}_1 - \hat{\tau}^{p}_1), \cdots\Big\}\Big).
\end{split}
\end{equation}
We design the numerical version of decision function $\tilde{d}_p$ based on the logical operations of the decision function $d_p$ as follows:
Given two inputs $A, B \in \{0, 1\}$, $A$ \texttt{AND} $B$, $A$ \texttt{OR} $B$, and \texttt{NOT} $A$ are substituted with  $A\cdot{B}$, $1 - (1-A)\cdot{(1-B})$, and $1-A$, respectively.

Using the predictions $\tilde{Y}^p = \{\tilde{y}^p_0, \tilde{y}^p_1, \cdots\}$ and the ground truth $Y^p = \{{y}^p_0, {y}^p_1, \cdots\}$, we calculate precision and recall of our predictions.
We further compute the loss $\mathcal{L}$ to optimize the parameters $\hat{T}_p$, which will be explained in Subsection~\ref{subsubsec:3_2_3_penalty_method}.

\noindent \textbf{Backward Pass.} 
As all the operations except HSF are differentiable, we define a surrogate gradient $\Theta'$ for $z_{ij} = \hat{q}_{ij} - \hat{\tau}^{p}_i$ as follows:
\begin{equation}
    \Theta'(z_{ij}) = \begin{cases}
            0, & \text{if $|z_{ij}| > w_i$,} \\
            \frac{\partial}{\partial z_{ij}} \Big(\frac{1}{2} \sin({\frac{\pi z_{ij}}{2w_i}}) +  \frac{1}{2}\Big)
            = {\frac{\pi}{4w_i}}{\cos({\frac{\pi z_{ij}}{2w_i}})},
            &\text{otherwise,}
        \end{cases}
\end{equation}
Note that defining the surrogate gradient as the above is same as approximating the step function HSF as follows:
\begin{equation}
    \widetilde{\text{HSF}}(z_{ij}) = \begin{cases}
            0, & \text{if $z_{ij} < -w_i$,} \\
            \frac{1}{2} \sin({\frac{\pi z_{ij}}{2w_i}}) +  \frac{1}{2} , &\text{if $-w_i \leq z_{ij} \leq w_i$,} \\
            1, & \text{otherwise,}.
        \end{cases}
\end{equation}
The above formulation is motivated from the idea that the threshold should be affected by samples near the threshold.
The width $w_i$ determines the amount of samples to be truncated.
We parameterize the width $w_i = \text{sigmoid}(\omega_i)$ so that it lies between $0$ and $1$.
We use the sine function to make the scores closer to the threshold have a larger effect during optimization.

Given the loss $\mathcal{L}$, the gradients with respect to the thresholds are computed using the surrogate gradient as follows:
\begin{equation}
\begin{split}
    \frac{\partial \mathcal{L}}{\partial \tau^{p}_i} 
    &= \sum_j\frac{\partial \mathcal{L}}{\partial \tilde{y}^p_j}\frac{\partial \tilde{y}^p_j}{\partial \tau^{p}_i} = \sum_j\frac{\partial \mathcal{L}}{\partial \tilde{y}^p_j}\frac{\partial \tilde{y}^p_j}{\partial \text{HSF}(z_{ij})}\frac{\partial \text{HSF}(z_{ij})}{\partial \tau^{p}_i} \\
    &= \sum_j\frac{\partial \mathcal{L}}{\partial \tilde{y}^p_j}\frac{\partial \tilde{y}^p_j}{\partial \text{HSF}(z_{ij})}\frac{\partial \text{HSF}(z_{ij})}{\partial z_{ij}}\frac{\partial z_{ij}}{\partial \tau^{p}_i} \\
    &\approx \sum_j-\frac{\partial \mathcal{L}}{\partial \tilde{y}^p_j}\frac{\partial \tilde{y}^p_j}{\partial \text{HSF}(z_{ij})}\Theta'(z_{ij}).
\end{split}
\label{eq:threshold_partial} 
\end{equation}

While updating the thresholds, we also update the width $w_i$ using a surrogate gradient $\Theta'(w_i)$:
\begin{equation}
    \Theta'(w_i) = \begin{cases}
            0, & \text{if $|z_{ij}| > w_i$,} \\
            \frac{\partial}{\partial w_i} \Big(\frac{1}{2} \sin({\frac{\pi z_{ij}}{2w_i}}) +  \frac{1}{2}\Big)
            = -\frac{\pi z_{ij}}{4w^2_i}\cos({\frac{\pi z_{ij}}{2w_i}}),
            &\text{otherwise,}
        \end{cases}
\label{eq:width_surrogate_gradient} 
\end{equation}
Similar to Equation~\ref{eq:width_surrogate_gradient}, the final gradients with respect to the widths are computed as follows.
\begin{equation}
    \frac{\partial \mathcal{L}}{\partial w_i} 
    \approx \sum_j\frac{\partial \mathcal{L}}{\partial \tilde{y}^p_j}\frac{\partial \tilde{y}^p_j}{\partial \text{HSF}(z_{ij})}\Theta'(w_i)
\label{eq:width_partial} 
\end{equation}
We found that making the widths learnable stabilizes training by dynamically adjusting the window size depending on each subtask.

\noindent
\subsubsection{Penalty method}\label{subsubsec:3_2_3_penalty_method} 
To solve the maximization problem (maximizing recall) with constraints (target precision), we design a relaxed loss function $\mathcal{L}$ as follows:
\begin{equation}
\begin{split}
    \mathcal{L} = -\text{recall} + \alpha\max(\text{precision}_t - \text{precision}, 0).
\label{eq:penalty}
\end{split}
\end{equation}
$\max(\text{precision}_t - \text{precision}, 0)$ is a penalty term to ensure that the precision meets the target precision where $\alpha$ controls the strictness of the constraint.
In our experiments, we set $\alpha \gg 1$.

%% file: figures/4_method.tex
\begin{figure}[t] 
\begin{center}
\includegraphics[width=0.85\columnwidth]{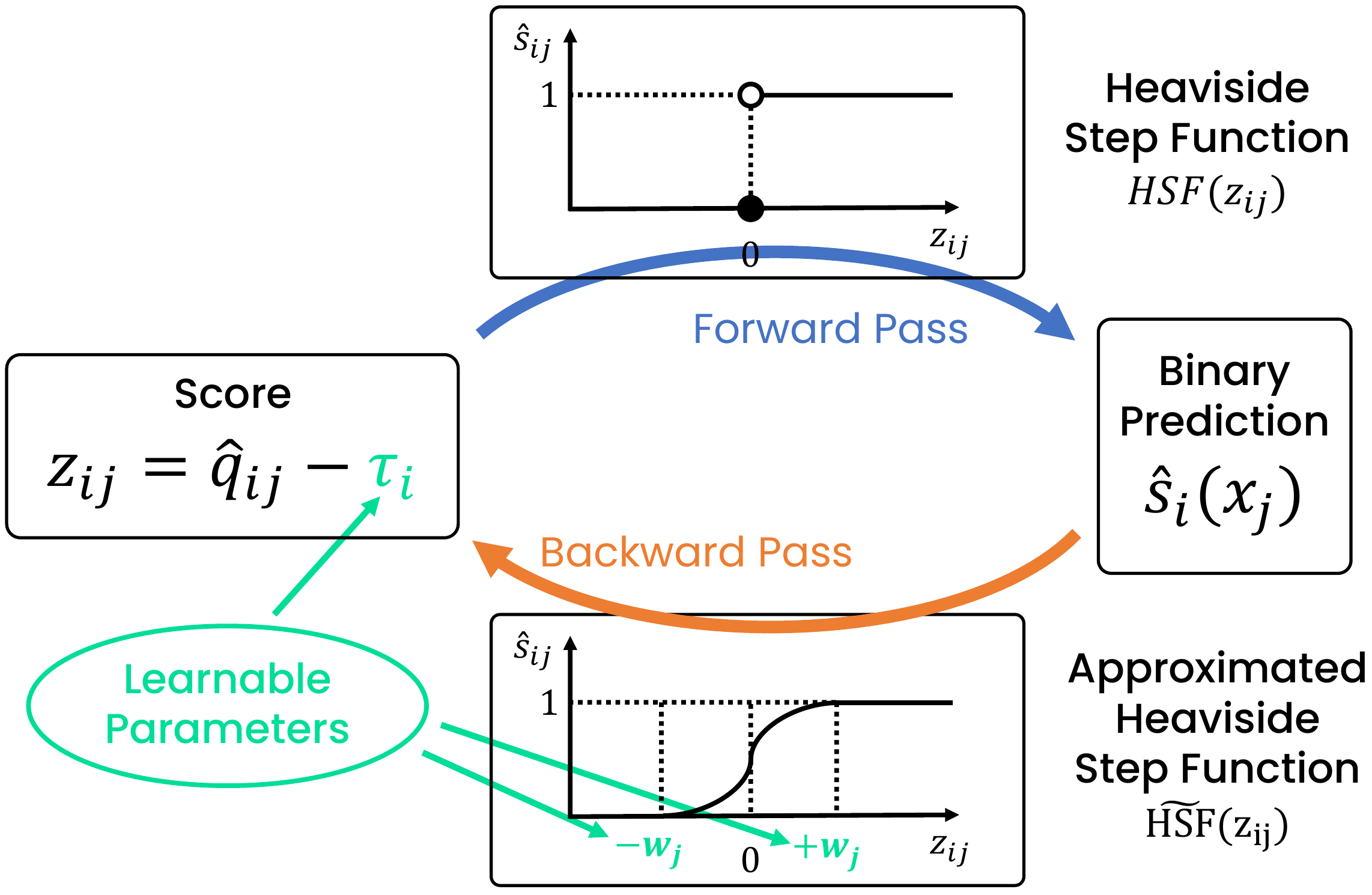}
\end{center}
\caption{
Surrogate Gradient Learning with Width Truncation for our proposed threshold optimization method, TruSThresh.
}
\vspace*{-1.0em}
\label{fig:4_method}
\end{figure}

%% file: sections/4_experiments.tex
\section{Experiments}\label{sec:4_experiments}
\subsection{Experiments on Moderation Use Cases}\label{subsec:4_1_moderation_exp}
\subsubsection{Settings.}
We show the performance of various threshold optimization methods on the two moderation use cases (Figure~\ref{fig:2_content_moderation_use_cases}).
In the first case, the ML model is used to filter out harmful content automatically.
The second case covers where the model is utilized to skip harmless content automatically.
We compare our method with other baselines in real-world moderation scenarios above.
Both scenarios utilize $n$ subtasks to classify whether inappropriate elements are included in the given content (class $1$) or not (class $0$).
Then, we can define decision functions for the two cases as follows:

\textit{Use case 1.}
\begin{equation}\label{eq:4_exp_1_use_case_1}
    d(O_{S}(x)) = \Big(s_{1}(x) \; \texttt{OR} \; s_{2}(x) \; \texttt{OR} \; \cdots \; \texttt{OR} \; s_{n}(x)\Big)
\end{equation}

\textit{Use case 2.}
\begin{equation}\label{eq:4_exp_1_use_case_2}
    \begin{split}
    d(O_{S}(x)) = \Big((\texttt{NOT}\; s_{1}(x)) \; \texttt{AND} \; (\texttt{NOT}\; s_{2}(x)) \; &\texttt{AND} \; \\
    \cdots \; &\texttt{AND} \; (\texttt{NOT}\; s_{n}(x))\Big)
    \end{split}
\end{equation}

As we concentrate on the case where we guarantee a reliable moderation decision, i.e., exceed the given target precision, we ignore the case where the precision from the optimized threshold does not surpass the target precision.\footnote{The implementation of experiments is publicly released at \url{https://github.com/hyperconnect/trusthresh}.}

\subsubsection{Dataset.}
We design the experiments to mimic the real-world moderation scenarios by utilizing \textbf{UnSmile}~\cite{kang2022korean}, a Korean hate speech dataset.
UnSmile contains $15,005$/$3,737$ training/test samples of $n=9$ hate speech subtasks. 
We use pretrained model released by the authors to obtain the prediction scores of the validation set and construct the policy-dependent dataset $\mathcal{D}_p$.
\input{tables/3_sota_case_1}

\subsubsection{Baselines.}
For our method, \textbf{TruSThresh}, we set the hyperparameters $\tau=0.5$, $w=0.1$, and learning rate $0.01$ to optimize for $1000$ iterations.
We compare our method with the following baselines:
\textbf{defThresh}~\cite{pellegrini2021fast} uses a single value $\tau$ for all the thresholds, where we use $\tau=0.5$ in the experiment.
\textbf{greedyThresh} calculates recall at precision for each subtask on evenly spaced threshold range between $0$ and $1$ to iteratively select the best thresholds for each subtask.
\textbf{SGLThresh} \cite{pellegrini2021fast}, similar to ours, also employs surrogate gradient of HSF to optimize the thresholds by the gradient descent.
It uses $sigmoid(\sigma_i z)$ to calculate the surrogate gradient of the step function $\mathds{1}(z > \tau_i)$ where $\sigma_i$ and $\tau_i$ is learnable.
We set $\tau=0.3$, $\sigma=50$, and learning rate $0.001$ to optimize for $4000$ iterations.
\textbf{Bespoke} represents $f_p(x)$ in Equation \ref{eq:use_case_1} and \ref{eq:use_case_2}, i.e., a model trained specifically for the target policy, serving as a strong baseline.

\subsubsection{Results.}
Table \ref{tab:sota_case_1} demonstrates the superior performance of our method compared to the state-of-the-art.
Especially, our method shows better performance compared to Bespoke in the second use case, even though it only took a few seconds to optimize for the thresholds.
In contrast, defThresh often fails to reach the target precision, and greedyThresh shows limited recall.

SGLThresh successfully reaches high precision in the first case but shows limited performance with higher precisions in the second case.
We suspect the second case is a more challenging scenario, as it is composed of \texttt{AND} operations, significantly decreasing the number of positive samples.
Even though increasing the penalty weight $\alpha$ seems beneficial for the second case since it helps reach the target precision by being more strict on the precision constraint, SGLThresh yields a lower recall compared to ours.

We also observe that using larger $\alpha$ degrades the recall for both TruSThresh and SGLThresh although they meet the target precision.
In practice, one can increase $\alpha$ until it satisfies the precision constraint, where further increases will negatively impact the recall.

\subsection{Experiments on Real-World Content Moderation}
\input{figures/5_real_world}

\subsubsection{Settings}
To show the effectiveness of TruSThresh in the wild, we conduct further experiments on the real-world moderation data.
We collect $23,596$ moderation samples on the live social discovery platform \textbf{\textit{Azar}}, which has been downloaded over $500$ million times.
The samples are annotated by professional human moderators.
We compare TruSThresh with SGLThresh and heuristic thresholds set by domain experts.
These methods search the optimal thresholds of ten subtasks for a decision function designed by the domain experts based on the application's moderation policy.
Note that we cannot report confidential information such as the ratio between true/false classes, details of each subtask, and the exact recall value.

\subsubsection{Results}
Figure~\ref{fig:5_real_world} shows the experimental results on real-world moderation data.
TruSThresh achieves the best recall over all the target precisions compared to the baselines.
Especially, at precision $0.985$, TruSThresh relatively improves recall by 5.6\% and 6.5\% compared to SGLThresh and heuristic thresholds, respectively.
We emphasize that TruSThresh reduces a considerable amount of annotation cost by only tuning the thresholds.
Moreover, unlike TruSThresh, SGLThresh fails to meet higher target precisions ($>0.985$).
These results verify that TruSThresh is highly effective in real-world content moderation systems.

\subsection{Experiments on Different Target Metric}\label{subsec:4_2_f1}
\subsubsection{Settings.}
We follow the settings of \cite{pellegrini2021fast}, where it optimizes the thresholds to maximize micro-averaged F1 score.
The score is defined in the multi-label classification setting as follows:
\begin{equation}\label{eq:micro_f1}
    \text{Micro-averaged F1} = 2\dfrac{\sum_{c,n}^{}y_{n}^{c}\tilde{y}^{c}_{n}}{\sum_{c,n}y^{c}_{n} + \sum_{c,n}\tilde{y}_{n}^{c}}.
\end{equation}
$y^{c}_{n}$ and $\tilde{y}^{c}_{n}$ are the ground truth and the prediction, respectively, for class $c$ and instance $n$.

We compare the methods using the following task:
\begin{equation}
    \underset{T}{\text{maximize}}\text{ Micro-averaged F1}(T|\mathcal{D}),
\label{eq:22}
\end{equation}
where $T$ is a set of classwise thresholds and $\mathcal{D}$ is the target multi-label classification dataset.

\subsubsection{Dataset.}
\input{tables/2_statistics_of_other_dataset}
Similar to \cite{pellegrini2021fast}, we use \textbf{DCASE17}~\cite{Mesaros2019_TASLP} and \textbf{DCASE19}~\cite{serizel2018large}, which is the multi-labeled sound event detection dataset.
We extend the experiment to different modalities by adding \textbf{EurLex}~\cite{chalkidis2019large} and \textbf{CelebA}~\cite{liu2015faceattributes}, which is the multi-label text and image dataset, respectively.
We target the validation set if there exists no specific test set.
Otherwise, we combine validation and test set to build a single target dataset.
Table~\ref{tab:2_statistics_of_other_datasets} summarizes the statistics of each target dataset.

\subsubsection{Models.}
For DCASE17 and DCASE19, we use the same prediction
scores of \citet{pellegrini2021fast}.
For CelebA, we train ResNet18~\cite{He_2016_CVPR} for all the subtasks using the train set with resolution $64$ for $30$ epochs with learning rate $0.001$, weight decay $0.01$, and batch size $256$ via AdamW optimizer~\cite{loshchilov2018decoupled} with cosine learning rate decay~\cite{loshchilov2016sgdr}.
For EurLex, we fine-tune pretrained RoBERTa~\cite{liu2019roberta} using the default settings of HuggingFace~\cite{wolf2019huggingface} with batch size $32$ for $10000$ steps.

\subsubsection{Baselines.}
For our method, \textbf{TruSThresh}, we set the hyperparameters $\tau=0.5$, $w=0.04$, and learning rate $0.01$ to optimize the thresholds for $400$ iterations.
We choose defThresh, greedyThresh, and SGLThresh as the baselines.
For greedyThresh, we select the thresholds that maximizes F1 for each subtask.
SGLThresh is originally designed for maximizing micro-averaged F1 score, hence being the primary competitor in this setting.
We use the default hyperparameters of SGLThresh, using $\tau=0.3$, $\sigma=50$, learning rate $0.01$, and $100$ iterations, except for DCASE2019 where we use $\tau=0.5$ to follow the original implementation \cite{pellegrini2021fast}.
For CelebA and EurLex, we optimize for $1000$ and $250$ iterations, respectively.

\subsubsection{Results.}
\input{tables/5_sota_f1}
Table \ref{tab:sota_f1} shows that our method shows better or comparable results than other baselines across different datasets of various modalities, even though our method is not explicitly designed to optimize the F1 score.
The results imply that our method can be further extended for other ranking-based metrics.
Especially, our method wins over SGLThresh \cite{pellegrini2021fast} on DCASE17 and DCASE19, which is the dataset that they used to target the F1 score.
Additionally, greedyThresh shows better performance than SGLThresh on DCASE17, serving as an effective baseline albeit its simplicity.
However, its performance drops on EurLex, and we suspect that it is due to the complexity of the problem; EurLex contains a large number of classes compared to others.

%% file: tables/3_sota_case_1.tex
\begin{table}[t]
\caption{
Experiments on two moderation use cases (Figure~\ref{fig:2_content_moderation_use_cases}).
``-'' is the case where it fails to meet the target precision.
Highest recall for each target precision is highlighted in bold.
}
\label{tab:sota_case_1}
\centering
\small
\begin{tabular}{c|l|c|c}

\toprule

Target & \multicolumn{1}{c|}{\multirow{2}{*}{Method}} & Recall in & Recall in \\
Precision & & Use case 1 & Use case 2 \\

\midrule
\multirow{9}{*}{0.9}
& Bespoke & 0.9069 & 0.2759 \\
\cline{2-4}
& defThresh $(\tau=0.5)$ & 0.9068 & - \\
& greedyThresh & 0.6734 & - \\
& SGLThresh $(\alpha=8)$ & 0.9422 & - \\
& SGLThresh $(\alpha=32)$ & 0.9425 & 0.2246 \\
& SGLThresh $(\alpha=128)$ & 0.9422 & 0.2171\\
\cline{2-4}
& TruSThresh $(\alpha=8)$ & \textbf{0.9443} & \textbf{0.3080} \\
& TruSThresh $(\alpha=32)$ & 0.9411 & 0.2631\\
& TruSThresh $(\alpha=128)$ & 0.9372 & 0.2599\\

\midrule
\multirow{9}{*}{0.95}
& Bespoke & 0.8694 & 0.1679 \\
\cline{2-4}
& defThresh $(\tau=0.5)$ & - & -\\
& greedyThresh & 0.4882 & - \\
& SGLThresh $(\alpha=8)$ & 0.8451 & -\\
& SGLThresh $(\alpha=32)$ & 0.8455 & 0.0246\\
& SGLThresh $(\alpha=128)$ & 0.8440 & 0.0631\\
\cline{2-4}
& TruSThresh $(\alpha=8)$ & \textbf{0.8512} & 0.1668 \\
& TruSThresh $(\alpha=32)$ & 0.8490 & \textbf{0.1679} \\
& TruSThresh $(\alpha=128)$ & 0.8440 & 0.1412 \\

\midrule
\multirow{9}{*}{0.975}
& Bespoke & 0.7855 & 0.0235 \\
\cline{2-4}
& defThresh $(\tau=0.5)$ & - & - \\
& greedyThresh & 0.3216 & - \\
& SGLThresh $(\alpha=8)$ & - & - \\
& SGLThresh $(\alpha=32)$ & 0.7145 & 0.0096 \\
& SGLThresh $(\alpha=128)$ & 0.7077 & 0.0481 \\
\cline{2-4}
& TruSThresh $(\alpha=8)$ & - & \textbf{0.1444} \\
& TruSThresh $(\alpha=32)$ & \textbf{0.7448} & 0.1348\\
& TruSThresh $(\alpha=128)$ & 0.7277 & 0.0995 \\

\bottomrule
\end{tabular}
\end{table}

%% file: figures/5_real_world.tex
\begin{figure}[t] 
\begin{center}
\includegraphics[width=0.85\columnwidth]{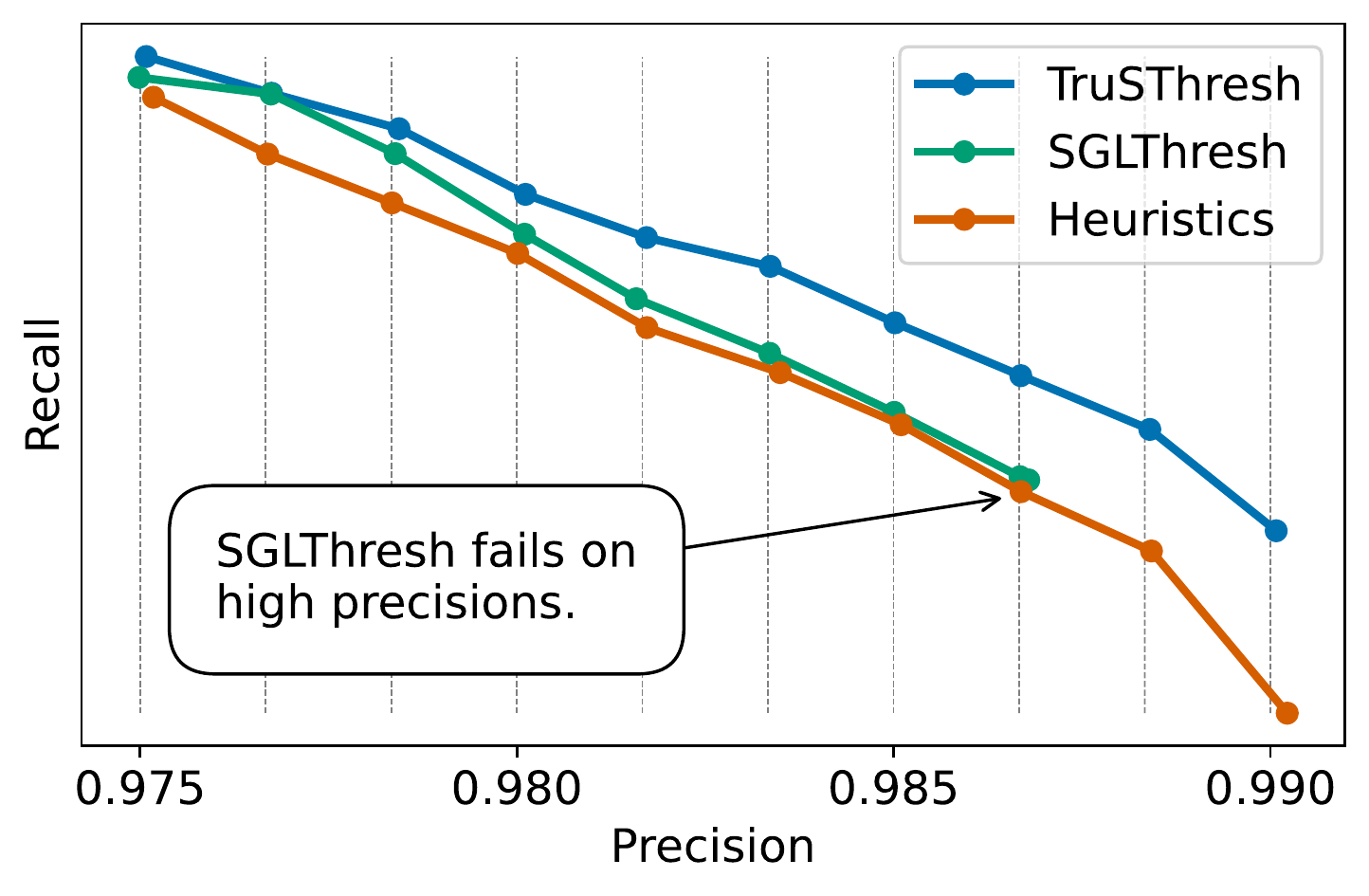}
\end{center}
\vspace*{-1em}
\caption{
Precision-recall curve over high precision.
Dashed lines indicate the target precision.
}
\vspace*{-1em}
\label{fig:5_real_world}
\end{figure}

%% file: tables/2_statistics_of_other_dataset.tex
\begin{table}[t]
\caption{Dataset used for \textit{micro-averaged F1 score.}}
\footnotesize
\vspace*{-1em}
\label{tab:2_statistics_of_other_datasets}
\begin{center}
\begin{tabular}{lrr}
\toprule
Dataset & \# of Classes & \ \# of Samples \\
\midrule
DCASE2017 & 17 & 1,591 \\
DCASE2019 & 10 & 1,814 \\
CelebA & 40 & 39,829 \\
EurLex & 746 & 11,920 \\
\bottomrule
\end{tabular}
\end{center}
\vspace*{-1em}
\end{table}

%% file: tables/5_sota_f1.tex
\begin{table}[t]
\caption{Experiments on Micro-averaged F1 Score.}
\vspace*{-1em}
\footnotesize
\label{tab:sota_f1}
\centering
\begin{tabular}{c|cccc}

\toprule

Method & DCASE17 & DCASE19 & CelebA & EurLex \\

\midrule
defThresh & 0.566 & 0.706 & 0.772 & 0.564 \\
greedyThresh & 0.636 & 0.728 & 0.771 & 0.593 \\
SGLThresh & 0.627 & 0.731 & \textbf{0.783} & \textbf{0.641} \\
TruSThresh & \textbf{0.641} & \textbf{0.732} & \textbf{0.783} & 0.639\\




\bottomrule
\end{tabular}
\vspace*{-5mm}
\end{table}

%% file: sections/5_analysis.tex
\section{Analysis}\label{sec:5_analysis}

\subsection{Ablation Study}\label{subsec:5_1_ablation_study}
\input{tables/6_ablation}
As mentioned in Subsection \ref{subsec:3_2_our_approach}, our method contains two key components: Score Normalization (SN) and Width Tuning (WT).
In Table \ref{tab:ablation}, we observe the performance when we remove each component.
We use the strictness of $\alpha=32$ for all the experiments.
We verify the effectiveness of both components, where SN greatly helps precisely target the precision.
As a result, recall increases while satisfying the precision constraint.

\subsection{Comparison with SGLThresh}\label{subsec:5_2_comparison}
\input{tables/7_sgl_ablation}
\noindent \textbf{Applying Score Normalization to SGLThresh.}
To the best of our knowledge, SGLThresh \cite{pellegrini2021fast} is the first work that applied surrogate gradient learning for threshold optimization.
To show the capability of our SN component, we attach the component to SGLThresh, comparing the downstream performance.
Original SGLThresh does not contain any score normalization involved.
We use the same settings of Subsection \ref{subsec:5_1_ablation_study}.
Table \ref{tab:sgl_ablation} demonstrates that our SN component is effective, similar to Subsection \ref{subsec:5_1_ablation_study}. 
SN helps increase the recall by setting the precision to be closer to the target precision.

\subsection{Optimization Analysis} \label{subsec:5_3_optimization}
The above subsection shows that the performance of SGLThresh is inferior to that of TruSThresh even with score normalization.
To further investigate the difference between the two methods, we analyze how learnable parameters are optimized during training.
The main difference comes from dissimilar approximations of HSF.
Given the thresholded prediction $z=q-\tau$, our method uses the sinusoidal function with a learnable width size, where SGLThresh uses $sigmoid(\sigma z)$ with learnable $\sigma$.
Both methods control the spread that depends on each subtask via classwise $w$ and $\sigma$, respectively.

\input{figures/5_analysis}
To observe whether each parameter is sufficiently being controlled to match the characteristics of each subtask, we plot how subtask-specific parameters are changing, i.e., plot $\sigma_{T}$ and $w_{T}$ where $T$ is the number of optimization steps.
We use the same settings from Subsection \ref{subsec:4_1_moderation_exp}, where each method has to optimize nine parameters per subtask.
We set the initial $\sigma_{T=0}=\{10, 25, 50\}$ and width $w_{T=0}=0.1$.
We also test smaller $\sigma$ than the original implementation of SGLThresh, which uses $\sigma=50$, to check whether SGLThresh successfully increases $\sigma$.
To fairly compare different methods, we compare the percentage difference from the initially set parameter, i.e., $\sigma_{T=t}/\sigma_{T=0}$ and $w_{T=t}/w_{T=0}$, at iteration step $t$.

Figure~\ref{fig:4_analysis} (a) shows that the classwise $\sigma$ learned by SGLThresh stays constant after few iterations.
Further, $\sigma$ is not much different between tasks, especially for $\sigma=50$, the original setting of SGLThresh.
However, after applying SN, it diversifies the thresholds, undergoing more extensive changes, as shown in Figure~\ref{fig:4_analysis} (b).
Unlike SGLThresh, we can observe in Figure~\ref{fig:4_analysis} (c) that the width of our TruSTHresh quickly changes.
It seems likely that they are being actively customized to each subtask score distribution.

We can analyze the above behavior more clearly by revisiting the partial derivatives of SGLThresh:
\begin{equation}
    {\partial \widetilde{\text{HSF}}_{\text{SGLThresh}}(z)}/{\partial \sigma} = z \cdot \text{sigmoid}(\sigma z) \cdot \text{sigmoid}(-\sigma z).
\label{eq:sigmoid_derivative}
\end{equation}
The above equation yields a small maximum value around $0.0045$ for the initial $\sigma_{T=0}=50$.
We suspect that the gradient of SGLThresh for each $\sigma$ vanishes, while our approximated HSF avoids this problem.

Finally, we compare the optimization speed in Figure \ref{fig:4_analysis} (d).
Even though the speed of SGLThresh significantly improves after applying score normalization, TruSTHresh is the quickest to converge, having the computational upper hand compared to SGLThresh.
We suspect that the design of subtask-specific width and the unimportant samples' gradients being truncated makes our method quickly optimize to the optimal thresholds with less number of steps.

%% file: tables/6_ablation.tex
\begin{table}[t]
\caption{Ablation study on Score Normalization and Width Tuning.
Experiments are conducted on \textit{Use case 2} with $\alpha=32$.}
\footnotesize
\label{tab:ablation}
\centering
\begin{tabular}{c|cc|cc}

\toprule

Target & Score & Width & \multirow{2}{*}{Precision} & \multirow{2}{*}{Recall} \\
Precision & Normalization & Tuning & & \\

\midrule
\multirow{4}{*}{0.9}
& \xmark & \xmark & 0.9286 & 0.1807 \\
& \xmark & \vmark & 0.9272 & 0.1497 \\
& \vmark & \xmark & 0.9046 & 0.2535 \\
& \vmark & \vmark & 0.9011 & \textbf{0.2631} \\

\midrule
\multirow{4}{*}{0.95}
& \xmark & \xmark & 0.9737 & 0.0396 \\
& \xmark & \vmark & 0.9697 & 0.0342 \\
& \vmark & \xmark & 0.9506 & 0.1647 \\
& \vmark & \vmark & 0.9518 & \textbf{0.1679} \\

\midrule
\multirow{4}{*}{0.975}
& \xmark & \xmark & 1.0000 & 0.0000 \\
& \xmark & \vmark & 1.0000 & 0.0000 \\
& \vmark & \xmark & 0.9810 & 0.1102 \\
& \vmark & \vmark & 0.9767 & \textbf{0.1348} \\

\bottomrule
\end{tabular}
\end{table}

%% file: tables/7_sgl_ablation.tex
\begin{table}[t]
\caption{
Comparing SGLThresh with TruSThresh.
Experiments are conducted on \textit{Use case 2} with $\alpha=32$.
For SGLThresh+SN, we apply score normalization to SGLThresh.
}
\label{tab:sgl_ablation}
\centering
\footnotesize
\begin{tabular}{c|c|cc}

\toprule

Target & \multirow{2}{*}{Method} & \multirow{2}{*}{Precision} & \multirow{2}{*}{Recall} \\
Precision & & & \\

\midrule
\multirow{3}{*}{0.9}
& SGLThresh & 0.9013 & 0.2246 \\
& SGLThresh+SN & 0.9026 & 0.2578 \\
& TruSThresh & 0.9011 & \textbf{0.2631} \\

\midrule
\multirow{3}{*}{0.95}
& SGLThresh & 0.9583 & 0.0246 \\
& SGLThresh+SN & 0.9517 & 0.1476 \\
& TruSThresh & 0.9518 & \textbf{0.1679} \\

\midrule
\multirow{3}{*}{0.975}
& SGLThresh & 1.0000 & 0.0096 \\
& SGLThresh+SN & 0.9828 & 0.1219 \\
& TruSThresh & 0.9767 & \textbf{0.1348} \\

\bottomrule
\end{tabular}
\vspace*{-1em}
\end{table}

%% file: figures/5_analysis.tex
\begin{figure}
    \centering
    \subfloat[Parameter Changes of SGLThresh]{\includegraphics[width=0.49\columnwidth]{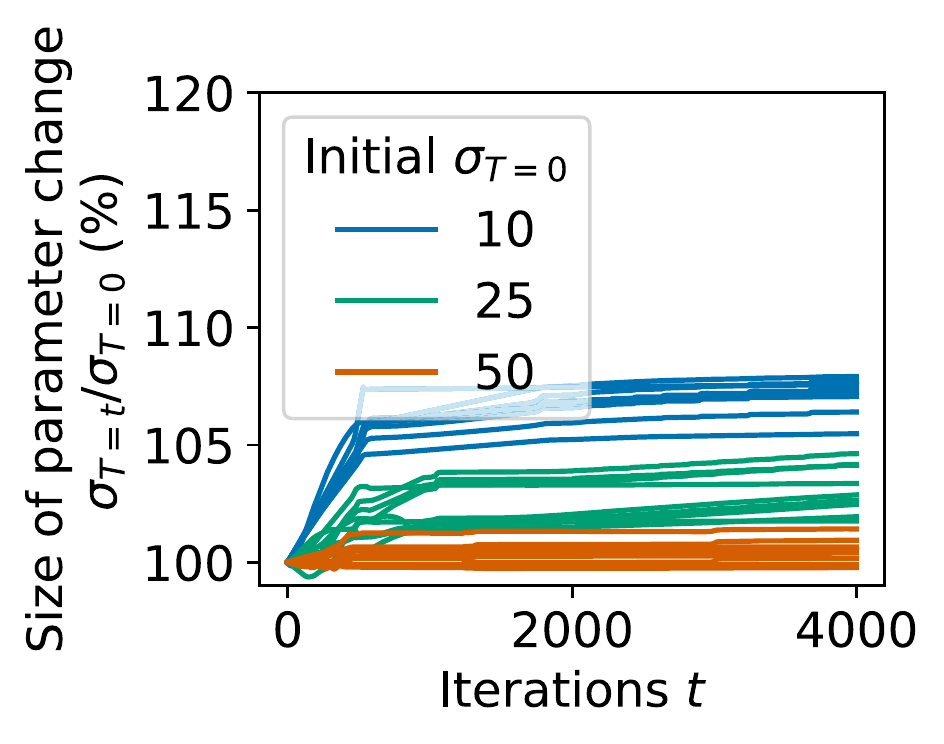} \hspace{0.3em}}
    \subfloat[Parameter Changes of SGLThresh + Score Normalization]{\includegraphics[width=0.49\columnwidth]{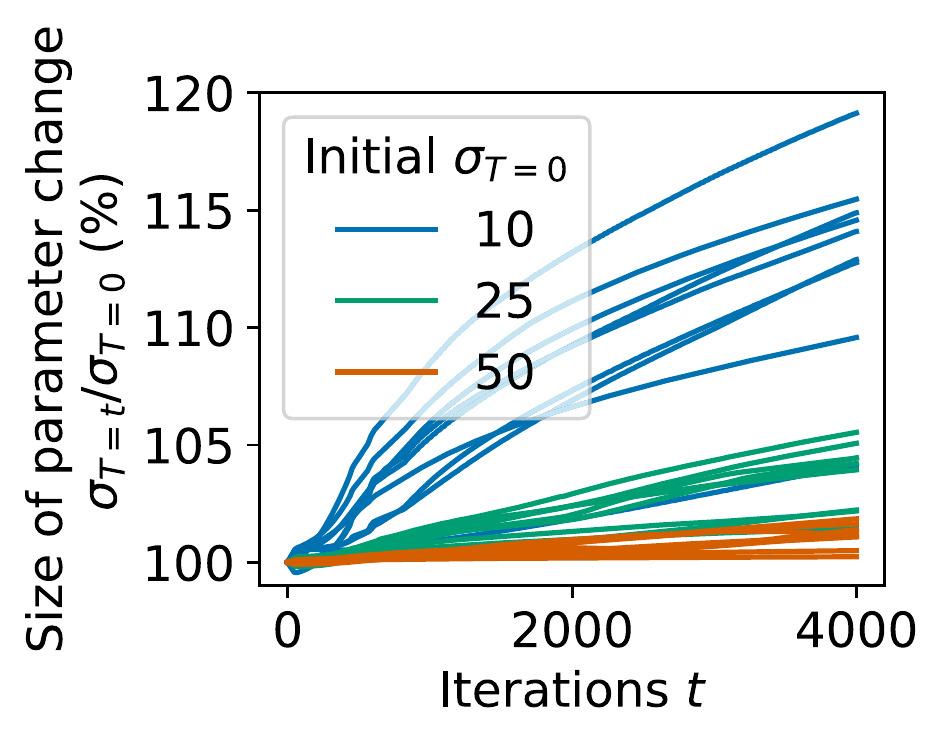}}\\
    \subfloat[Parameter Changes of TruSThresh]{\includegraphics[width=0.49\columnwidth]{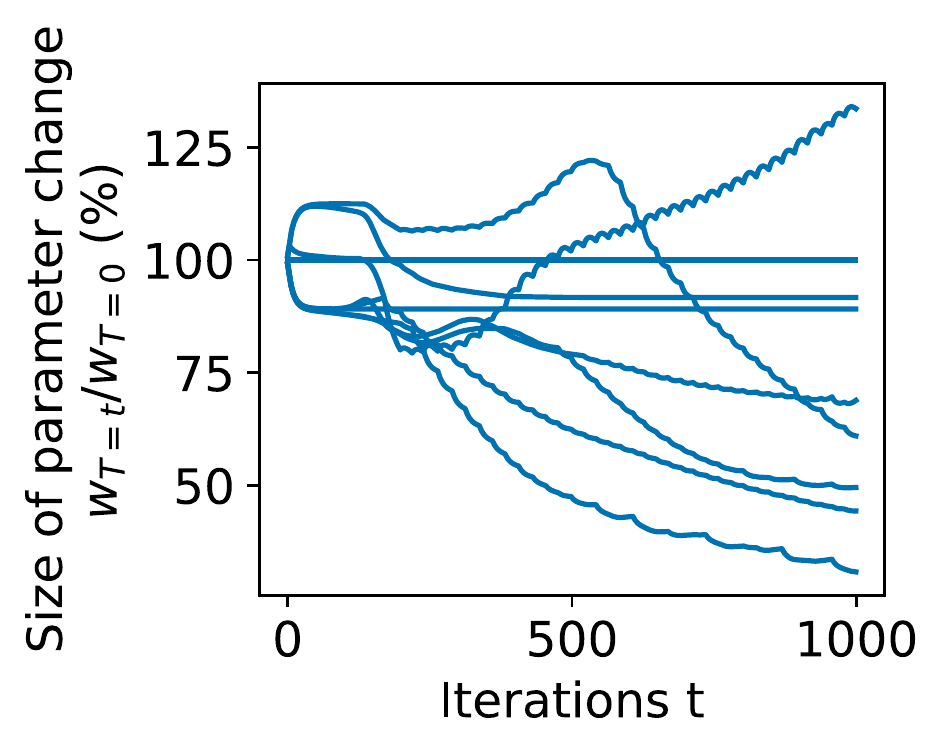} \hspace{0.3em}}
    \subfloat[Convergence Speed Comparison]{\includegraphics[width=0.49\columnwidth]{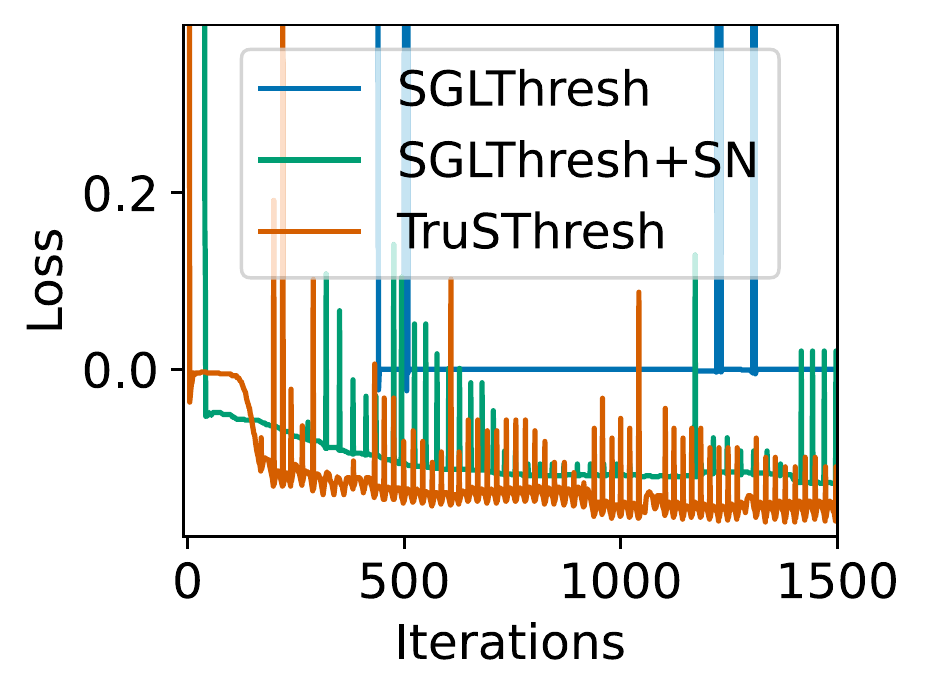}}

    \caption{
        Method behaviors during optimization.
    }
\vspace*{-2.0em}
    \label{fig:4_analysis}
\end{figure}

%% file: sections/6_related_work.tex
\section{Related Work}\label{sec:7_related_work}
\subsection{Machine Learning for Content Moderation}\label{subsec:7_1_machine_learning_model_based_content_moderation}
ML models have been widely adopted for content moderation in the last decade.
Since it is very difficult to obtain data on content moderation due to the sensitivity of harmful content, studies on content moderation are mainly conducted in industry rather than academia.
In the industry, research on collaboration between ML models and human moderators is predominant~\cite{green2019principles,lubars2019ask,markov2022holistic}, rather than directly revealing the ML models operating on social media platforms.
Studies conducted in academia mainly focus on text content that is easily available online.
Hate speech detection is one of the popular research topics on content moderation.
Using benchmark datasets obtained from Twitter~\cite{waseem2016hateful,davidson2017automated,founta2018large}, Yahoo!~\cite{nobata2016abusive}, and Reddit~\cite{qian2019benchmark}, hate speech detection models~\cite{djuric2015hate,badjatiya2017deep} have been proposed based on the state-of-the-art architecture such as convolutional neural networks~\cite{park2017one}, recurrent neural networks~\cite{founta2019unified}, and Transformer~\cite{elsherief2021latent}.
Several studies~\cite{chang2018deep,chang2021killing} have proposed spoiler detection models to protect users from spoilers that ruin the pleasures of the users for the creative works.
Recently, a detection model~\cite{hu2021detection} tracking harmful content that sells illicit drugs on Instagram has been proposed based on deep multi-modal multi-label learning.

\subsection{Threshold Optimization}\label{subsec:7_2_threshold_optimization}
Several research areas are loosely related to our study in terms of finding the optimal thresholds that maximize the target metrics.
When we formulate the threshold optimization problem as minimizing a real-valued function output, we can consider it as the unconstrained nonlinear optimization problem~\cite{nelder1965simplex,nocedal2006line}.
Also, the algorithm configuration problem~\cite{hutter2011sequential} concentrates on the variant of the above situation where yielding the output is expensive.
However, calculating the metrics such as the F1 score is a cheap operation compared to the targets of the algorithm configuration problem.
Even though our work did not concentrate on improving the model itself, there are recent efforts that directly optimize the model to the black-box metrics.
\citet{jiang2020optimizing} estimate the gradients of the metrics to utilize, and \citet{huang2019addressing} address the mismatch between the loss to train the model and the metric to evaluate, where they adjust the loss to follow the metric.
Furthermore, some concentrate on ranking-based metrics, such as precision or recall, similar to our work.
\citet{eban2017scalable} apply relaxation to the metrics to make it tractable, and \citet{revaud2019learning} and \citet{henderson2016end} utilize mean average precision loss to solve image retrieval or object detection problems.
The most similar work is the threshold optimization literature \cite{kong2020sound,cances2019evaluation,pellegrini2021fast}.
\citet{kong2020sound} directly optimize the F1 score via simple gradient descent with heuristically chosen step size.
\citet{cances2019evaluation} search the thresholds initially within a coarse range to iteratively reduce the search space.
Finally, SGLThresh~\cite{pellegrini2021fast} improve the previous two methods, where our method share the key idea of surrogate gradient learning.

%% file: sections/7_conclusion.tex
\section{Conclusion}\label{sec:7_conclusion}
In this study, we describe real-world content moderation scenarios based on the multiple subtask approach to cope with various moderation policies.
To make moderation decisions more reliable from multiple subtask predictions of ML models, we propose a threshold optimization method that finds the optimal thresholds for the subtasks.
Experiments on synthetic and real-world content moderation datasets show that our proposed method improves recall while preserving high precision, optimizing within seconds without any model parameter updates.
We believe our study will aid social media platforms that already operate ML-based content moderation systems or are considering building new systems.